# On the Application of Hierarchical Coevolutionary Genetic Algorithms: Recombination and Evaluation Partners

Uwe Aickelin
School of Computer Science
University of Nottingham
NG8 1BB   UK
uxa@cs.nott.ac.uk

Larry Bull
Faculty of Computing, Engineering & Mathematical Sciences
University of the West of England
Bristol BS16 1QY, U.K.

**Abstract**

This paper examines the use of a hierarchical coevolutionary genetic algorithm under different partnering strategies. Cascading clusters of sub-populations are built from the bottom up, with higher-level sub-populations optimising larger parts of the problem. Hence higher-level sub-populations potentially search a larger search space with a lower resolution whilst lower-level sub-populations search a smaller search space with a higher resolution. The effects of different partner selection schemes amongst the sub-populations on solution quality are examined for two constrained optimisation problems. We examine a number of recombination partnering strategies in the construction of higher-level individuals and a number of related schemes for evaluating sub-solutions. It is shown that partnering strategies that exploit problem-specific knowledge are superior and can counter inappropriate (sub-) fitness measurements.

## 1   INTRODUCTION

The use of coevolutionary computation for optimisation raises a number of new questions, one of which is addressed in this paper: the issue of intelligently selecting partners for both mating and evaluation from other evolving populations. This paper will look at a number of different partnering strategies when combined with a hierarchical scheme that uses a co-operative sub-population structure. We will evaluate the different strategies according to their optimisation performance on two constrained scheduling problems.

Within the hierarchical structure, all sub-populations follow different (sub-) fitness functions, so in effect they are only searching specific parts of the solution space. Following special crossover-operators, these parts are then gradually merged to full solutions. The advantage of such a divide and conquer approach is a reduced search space size and/or epistasis within the lower-level sub-populations which (potentially) makes the optimisation task easier for the genetic algorithm (GA) [Holland 1975].

The paper is arranged as follows: the following section describes the nurse scheduling and tenant selection problems. Pyramidal genetic algorithms and their application to these two problems are detailed in section 3. Section 4 explains the seven partnering strategies examined in the paper and section 5 describes their use and computational results. The final section discusses all findings and draws conclusions.

## 2   THE NURSE SCHEDULING AND TENANT SELECTION PROBLEMS

Two optimisation problems are considered in this paper, the nurse scheduling problem and the tenant selection problem. Both have a number of characteristics that make them an ideal testbed for the enhanced genetic algorithm using partnering strategies. Firstly, they are both in the class of NP-complete problems [e.g. Martello & Toth 1990]; hence, they are challenging problems. Secondly, they have proved resistant to optimisation by a standard genetic algorithm, with good solutions only found by using a novel strategy of indirectly optimising the problem with a decoder based genetic algorithm [Aickelin & Dowsland 2001]. Finally, both problems are similar multiple-choice allocation problems. For nurse scheduling, the choice is to allocate a shift-pattern to each nurse, whilst for the tenant selection it is to allocate an area of the mall to a shop. However, as the following more detailed explanation of the two will show, the two problems also have some very distinct characteristics making them different yet similar enough for an interesting comparison of results.

The nurse-scheduling problem is that of creating weekly schedules for wards of up to 30 nurses at a major UK hospital. These schedules have to satisfy working contracts and meet the demand for given numbers of nurses of different grades on each shift, whilst at the same time being seen to be fair by the staff concerned. The latter objective is achieved by meeting as many of the nurses' requests as possible and by considering historical information to ensure that unsatisfied requests and unpopular shifts are evenly distributed. Due to various hospital policies, a nurse can normally only work a sub-set of the in total 411 theoretically possible shift-patterns. For instance, a nurse should work either days or nights in a given week, but not both. The interested reader is directed to [Aickelin & Dowsland 2000] for further details of this problem.

For our purposes, the problem can be modelled as follows. Nurses are scheduled weekly on a ward basis such that they work a feasible pattern with regards to their contract and that the demand for all days and nights and for all qualification levels is covered. In total three qualification levels with corresponding demand exists. It is hospital policy that more qualified nurses are allowed to cover for less qualified one. Infeasible solutions with respect to cover are not acceptable. A solution to the problem would be a string, with the number of elements equal to the number of nurses. Each element would then indicate the shift-pattern worked by a particular nurse. Depending on the nurses' preferences, the recent history of patterns worked and the overall attractiveness of the pattern, a penalty cost is then allocated to each nurse-shift-pattern pair. These values were set in close consultation with the hospital and range from 0 (perfect) to 100 (unacceptable), with a bias to lower values. The sum of these values gives the quality of the schedule. 52 data sets are available, with an average problem size of 30 nurses per ward and up to 411 possible shift-patterns per nurse.

The problem can be formulated as an integer linear program as follows.

Indices:

$i = 1...n$ nurse index.

$j = 1...m$ shift pattern index.

$k = 1...14$ day and night index (1...7 are days and 8...14 are nights).

$s = 1...p$ grade index.

Decision variables:

$$x_{ij} = \begin{cases} 1 & \text{nurse } i \text{ works shift pattern } j \\ 0 & \text{else} \end{cases}$$

Parameters:

$n$ = Number of nurses.

$m$ = Number of shift patterns.

$p$ = Number of grades.

$$a_{jk} = \begin{cases} 1 & \text{shift pattern } j \text{ covers day / night } k \\ 0 & \text{else} \end{cases}$$

$$q_{is} = \begin{cases} 1 & \text{nurse } i \text{ is of grade } s \text{ or higher} \\ 0 & \text{else} \end{cases}$$

$p_{ij}$ = Preference cost of nurse $i$ working shift pattern $j$.

$N_i$ = Working shifts per week of nurse $i$ if night shifts are worked.

$D_i$ = Working shifts per week of nurse $i$ if day shifts are worked.

$B_i$ = Working shifts per week of nurse $i$ if both day and night shifts are worked.

$R_{ks}$ = Demand of nurses with grade $s$ on day respectively night $k$.

$F(i)$ = Set of feasible shift patterns for nurse $i$, where $F(i)$ is defined as

$$F(i) = \begin{cases} \sum_{k=1}^{7} a_{jk} = D_i & \forall j \in \text{day shifts} \\ or & \\ \sum_{k=8}^{14} a_{jk} = N_i & \forall j \in \text{night shifts} \\ or & \\ \sum_{k=1}^{14} a_{jk} = B_i & \forall j \in \text{combined shifts} \end{cases} \quad \forall i$$

Target function:

$$\sum_{i=1}^{n} \sum_{j \in F(i)} p_{ij} x_{ij} \quad \rightarrow \quad \text{min!}$$

Subject to:

1. Every nurse works exactly one feasible shift pattern:

$$\sum_{j \in F(i)} x_{ij} = 1 \quad \forall i \quad (1)$$

2. The demand for nurses is fulfilled for every grade on every day and night:

$$\sum_{j \in F(i)} \sum_{i=1}^{n} q_{is} a_{jk} x_{ij} \geq R_{ks} \quad \forall k, s \quad (2)$$

Constraint set (1) ensures that every nurse works exactly one shift pattern from his/her feasible set, and constraint set (2) ensures that the demand for nurses is covered for every grade on every day and night. Note that the definition of $q_{is}$ is such that higher graded nurses can substituted those at lower grades if necessary. Typical problem dimensions are 30 nurses of three grades and 411 shift patterns. Thus, the Integer Programming formulation has about 12000 binary variables and 100 constraints.

Finally for all decoders, the fitness of completed solutions has to be calculated. Unfortunately, feasibility cannot be guaranteed, as otherwise an unlimited supply of nurses, respectively overtime, would be necessary. This is a problem-specific issue and cannot be changed. Therefore, we still need a penalty function approach. Since the chosen encoding automatically satisfies constraint set (1) of the integer programming formulation, we can use the following formula, where $w_{demand}$ is the penalty weight, to calculate the fitness of solutions. Hence the penalty is proportional to the number of uncovered shifts and the fitness of a solution is calculated as follows.

$$\sum_{i=1}^{n} \sum_{j=1}^{m} p_{ij} x_{ij} + w_{demand} \sum_{k=1}^{14} \sum_{s=1}^{p} \max\left[ R_{ks} - \sum_{i=1}^{n} \sum_{j=1}^{m} q_{is} a_{jk} x_{ij}; 0 \right] \quad \rightarrow \quad \text{min!}$$

Here we use an encoding that follows directly from the Integer Programming formulation. Each individual represents a full one-week schedule, i.e. it is a string of *n* elements with *n* being the number of nurses. The *ith* element of the string is the index of the shift pattern worked by nurse *i*. For example, if we have 5 nurses, the string (1,17,56,67,3) represents the schedule in which nurse 1 works pattern 1, nurse 2 pattern 17 etc.

For comparison, all data sets were attempted using a standard Integer Programming package [Fuller 1998]. However, some remained unsolved after each being allowed 15 hours run-time on a Pentium II 200. Experiments with a number of descent methods using different neighbourhoods, and a standard simulated annealing implementation, were even less successful and frequently failed to find feasible solutions. A straightforward genetic algorithm approach failed to solve the problem [Aickelin & Dowsland 2000]. The best evolutionary results to date have been achieved with an indirect genetic approach employing a decoder function [Aickelin & Dowsland 2001]. However, we believe that there is further leverage in direct evolutionary approaches to this problem. Hence, we propose to use an enhanced pyramidal genetic algorithm in this paper.

The second problem is a mall layout and tenant selection problem; termed the mall problem here. The mall problem arises both in the planning phase of a new shopping centre and on completion when the type and number of shops occupying the mall has to be decided. To maximise revenue a good mixture of shops that is both heterogeneous and homogeneous has to be achieved. Due to the difficulty of obtaining real-life data because of confidentiality, the problem and data used in this research are constructed artificially, but closely modelled after the actual real-life problem as described for instance in Bean et al. [1988]. In the following, we will briefly outline our model.

The objective of the mall problem is to maximise the rent revenue of the mall. Although there is a small fixed rent per shop, a large part of a shop's rent depends on the sales revenue generated by it. Therefore, it is important to select the right number, size and type of tenants and to place them into the right locations to maximise revenue. As outlined in Bean et al. [1988], the rent of a shop depends on the following factors:

- The attractiveness of the area in which the shop is located.
- The total number of shops of the same type in the mall.
- The size of the shop.
- Possible synergy effects with neighbouring similar shops, i.e. shops in the same group (not used by Bean et al.).
- A fixed amount of rent based on the type of the shop and the area in which it is located.

This problem can be modelled as follows: Before placing shops, the mall is divided into a discrete number of locations, each big enough to hold the smallest shop size. Larger sizes can be created by placing a shop of the same type in adjacent locations. Hence, the problem is that of placing $i$ shop-types (e.g. menswear) into $j$ locations, where each shop-type can belong to one or more of $l$ groups (e.g. clothes shops) and each location is situated in one of $k$ areas. For each type of shop there will be a minimum, ideal and maximum number allowed in the mall, as consumers are drawn to a mall by a balance of variety and homogeneity of shops.

The size of shops is determined by how many locations they occupy within the same area. For the purpose of this study, shops are grouped into three size classes, namely small, medium and large, occupying one, two and three locations in one area of the mall respectively. For instance, if there are two locations to be filled with the same shop-type within one area, then this will be a shop of medium size. If there are five locations with the same shop-type assigned in the same area, then they will form one large and one medium shop etc. Usually, there will also be a maximum total number of small, medium and large shops allowed in the mall.

To test the robustness and performance of our algorithms thoroughly on this problem, 50 problem instances were created. All problem instances have 100 locations grouped into five areas. However, the sets differ in the number of shop-types available (between 50 and 20) and in the tightness of the constraints regarding the minimum and maximum number of shops of a certain type or size. Full details of the model, how the data was created, its dimensions and the differences between the sets can be found in [Aickelin 1999].

## 3    PYRAMIDAL GENETIC ALGORITHMS

Both problems failed to be optimised with a standard genetic algorithm [Aickelin & Dowsland 2000, 2001]. Our previous research showed that the difficulties were attributable to epistasis created by the constrained nature of the optimisation. Briefly, epistasis refers to the 'non-linearity' of the solution string [Davidor 1991], i.e. individual variable values which were good in their own right, e.g. a particular shift / location for a particular nurse / shop formed low quality solutions once combined. This effect was created by those constraints that could only be incorporated into the genetic algorithm via a penalty function approach. For instance, most nurses preferred working days; thus, partial solutions with many 'day' shift-patterns have a higher fitness. However, combining these shift-patterns leads to shortages at night and therefore infeasible solutions. The situation for the mall problem is similar yet more complex, as two types of constraints have to be dealt with: size constraints and number constraints.

In [Aickelin & Dowsland 2000] we presented a simple, and on its own unsuccessful, pyramidal genetic algorithm for the nurse-scheduling problem. A pyramidal approach can best be described as a hierarchical coevolutionary genetic algorithm where cascading clusters of sub-populations are built from bottom up, with higher-level sub-populations optimising larger parts of the problem. Thus, the hierarchy is not within one representation of the problem but rather between sub-populations which optimise different portions of the global problem. Hence, higher-level sub-populations search a larger search space with a lower resolution whilst lower-level sub-populations search a smaller search space with a higher resolution. A related hierarchical framework was presented using Genetic Programming [Koza 1991] whereby main program trees coevolve with successively lower level functions [e.g. Ahluwalia & Bull 1998]. The pyramidal GA can be applied to the nurse-scheduling problem in the following way:

- Solutions in sub-populations 1, 2 and 3 have their fitness based on cover and requests only for grade 1, 2 and 3 respectively.
- Solutions in sub-populations 4, 5 and 6 have their fitness based on cover and requests for grades (1+2), (2+3) and (3+1).
- Solutions in sub-population 7 optimise cover and requests for (1+2+3).
- Solutions in sub-population 8 solve the original (all) problem, i.e. cover for 1, for (1+2) and for (1+2+3).

The full structure is illustrated in figure 1. Sub-solution strings from lower populations are cascaded upwards using suitable crossover and selection mechanisms. For instance, fixed crossover points are used such that a solution from sub-population (1) combined with one from (1+2) forms a new solution in sub-population (1+2). Each sub-population performs 50% of crossovers uniform with two parents from itself. The other 50% are done by taking one parent from itself and the other from a suitable lower level population (picked at random) and then performing a fixed-point crossover. Bottom level sub-populations use only uniform crossover. The top level (all) population randomly chooses the second parent from all other populations. Although the full problem is as epistatic as before, the sub-problems are less so as the interaction between nurse grades is (partially) ignored. Compatibility problems of combining the parts are reduced by the pyramidal structure with its hierarchical and gradual combining.

Using this approach improved solution quality in comparison to a standard genetic algorithm was recorded. Initially roulette wheel selection based on fitness rank had been used to choose parents. The fitness of each sub-string is calculated using a substitute fitness measure based on the requests and cover as detailed above, i.e. the possibility of more qualified nurses covering for less-qualified ones is partially ignored. Unsatisfied constraints are still included via a penalty function. This paper will investigate various partnering strategies between the agents of the sub-populations to improve upon these results.

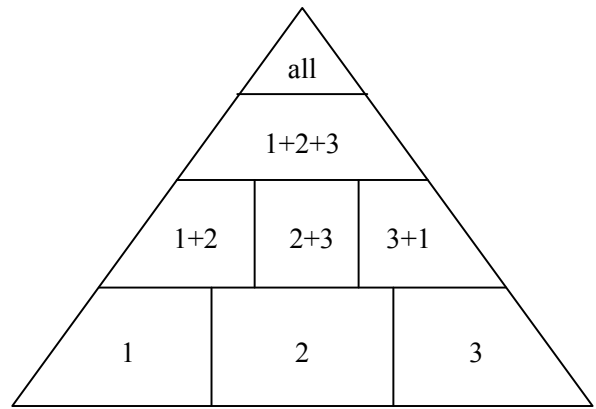

Figure 1: Nurse Problem Pyramidal Structure.

Similar to the nurse problem, a solution to the mall problem can be represented by a string with as many elements as locations in the mall. Each element then indicates what shop-type is to be located there. The mall is geographically split into different regions, for instance north, east, south, west and central. Some of the objectives are regional; e.g. the size of a shop, the synergy effects, the attractiveness of an area to a shop-type, whereas others are global, e.g. the total number of shops of a certain type or size.

The application of the pyramidal structure to the mall problem follows along similar lines to that of the nurse problem. In line with decomposing partitions into those with nurses of the same grade, the problem is now split into the areas of the mall. Thus, we will have sub-strings with all the shops in one area in them. These can then be combined to create larger 'parts' of the mall and finally full solutions.

However, the question arises how to calculate the substitute fitness measure of the partial strings. The solution chosen here will be a pseudo measure based on area dependant components only, i.e. global aspects are not taken into account when a substitute fitness for a partial string is calculated. Thus, sub-fitness will be a measure of the rent revenue created by parts of the mall, taking into account those constraints that are area based. All other constraints are ignored. A penalty function is used to account for unsatisfied constraints.

Due to the complexity of the fitness calculations and the limited overall population size, we refrained from using several levels in the hierarchical design as we did with the nurse scheduling. Instead a simpler two-level hierarchy is used as shown in figure 2: Five sub-populations optimising the five areas separately (1,2,3,4,5) and one main population optimising the original problem (all). Within the sub-populations 1-5 uniform crossover is used. The top-level population uses uniform crossover between two members of the population half the time and for the remainder a special crossover that selects one solution from a random sub-population that then performs a fixed-point crossover with a member of the top population.

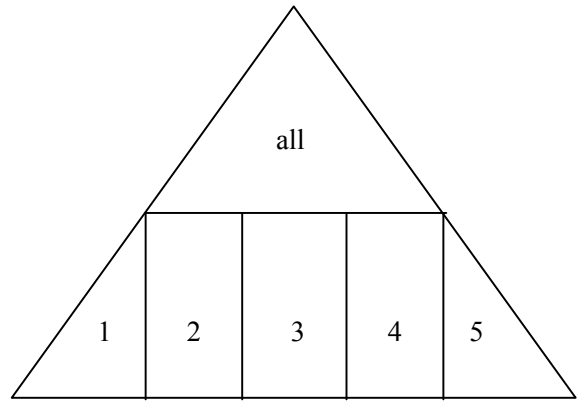

Figure 2: Mall Problem Pyramidal Structure.

The remainder of this paper will investigate ways to try to improve on previously found poor results by suggesting ways of combining partial strings more intelligently. An alternative, particularly for the mall problem, would be a more gradual build-up of sub-populations. Without increasing the overall population size, this would lead to more and hence smaller sub-populations. However, this more gradual approach might have enabled the algorithm to find good feasible solutions by more slowly joining together promising building blocks. This is in contrast to the relatively harsh two-level and three-level design where building blocks had to 'succeed' immediately. Exploring the exact benefits of a gradual build-up of sub-solutions would make for another challenging area of possible future research.

## 4    PARTNERING STRATEGIES

The problem of how to pick partners has been noted in both competitive and co-operative coevolutionary algorithms. Many strategies have been presented in the literature as summarised for instance in [Bull 1997]. In this paper, the following strategies are compared for their effectiveness in fighting epistasis in the pyramidal genetic algorithm optimising the nurse scheduling and the mall problems. As indicated, some strategies will be used for fitness evaluation, some for mating selection and some for both.

Rank-Selection (S) [Both]: This is the method used so far in our algorithms. Solutions are assigned a sub-fitness score based as closely as possible on the contribution of their partial string to full solutions. All solutions are then ranked within each sub-population and selection follows a roulette wheel scheme based on the ranks [e.g. Aickelin & Dowsland 2000].

Random (R) [Both]: Solutions choose their mating partners randomly from amongst all those in the sub-population their sub-population is paired with [e.g. Bull & Fogarty 1993].

Best (B) [Both]: In this strategy, each agent is paired with the current best solution of the other sub-population(s). In case of a tie, the solution with the lower population index is chosen [e.g. Potter & De Jong 1994].

Distributed (D) [Both]: Here each sub-population is spaced out evenly across a single toroidal grid [e.g. Ackley & Littman 1987]. Subsequently, solutions are paired with others on the same grid location in the appropriate other sub-populations. Children created in this way are inserted in an adjacent grid location. This is said to be beneficial to the search process because a consistent coevolutionary pressure emerges since all offspring appear in their parents' neighbourhoods and so there is potentially less variance in partners between generations [Husbands 1994]. In our algorithms, we use local mating with the neighbourhood set to the eight agents surrounding the chosen location.

Joined (J) [Mating only]: In nature, some species carry others internally with the relationship propagated from generation to generation. Thus, each individual represents a complete solution; i.e. all the parts have been joined together [e.g. Iba 1996]. In our case, this results in all sub-populations solving the original problem, i.e. we have a traditional parallel genetic algorithm. This means that all sub-populations use the full fitness function for evaluation and rank-proportional selection.

Attractiveness (A) [Mating only]: The five strategies described so far are general and do not make use of problem specific knowledge. However, there is a growing body of research [e.g. Stanley et al. 1994, Wolpert & Macready 1997], as well as our own previous work [Aickelin & Dowsland 2000], which suggests that approaches that exploit problem specific knowledge achieve better results. Here pairing is done as for the rank-selection strategy (S). However, the pair is only accepted with a probability proportional to their fitness or substitute fitness once combined. The probabilities are scaled such that if the (substitute) fitness $f_{comb}$ is equal or greater to the best-known fitness $f_{best}$ the pairing is automatically accepted. Otherwise the probability is $f_{comb} / f_{best}$ for the mall problem and the inverse for the nurse scheduling.

Partner Choice (C) [Mating only]: This approach again exploits problem specific knowledge and was inspired by an idea presented by Ronald [1995]. He solves Royal Roads and multi-objective optimisation problems using a genetic algorithm where the first parent is chosen following standard rules, i.e. proportional to its fitness. However, the second parent is not chosen according to its fitness, but depending on its 'attractiveness' to the first parent, which is measured on a different scale. Our approach will be slightly different. The first parent is still chosen according to its rank. But rather than picking one individual from the appropriate sub-population as the second parent, ten candidates are chosen at random. The second parent will then be chosen as the one that creates the fittest children with the first parent.

Best / Random (SR) [Evaluation only]: A solution is paired twice: with the best of the other sub-population(s) and with a random partner(s). The better of the two fitness values is recorded.

Rank-based / Random (SR) [Evaluation only]: A solution is paired twice: with roulette wheel selected solution(s) and with (a) random partner(s). The better of the two fitness values is recorded.

Random / Random (RR) [Evaluation only]: A solution is paired twice with random partner(s). The better of the two fitness values is recorded.

## 5    EXPERIMENTAL RESULTS

### 5.1    THE MODEL

To allow for fair comparison, the parameters and strategies used for both problems are kept as similar as possible. Both have a total population of 1000 individuals. These are split into sub-populations of size 100 for the lower-levels and a main population of size 300 for the nurse scheduling and respectively of size 500 for the mall problem. In principle, two types of crossover take place: within sub-populations, a two-parent-two-children parameterised uniform crossover with $p$=0.66 for genes coming from one parent takes place.

Each new solution created undergoes mutation with a 1% bit mutation probability, where a mutation would re-initialise the bit in the feasible range. The algorithm is run in generational mode to accommodate the sub-population structure better. In every generation, the worst 90% of parents of all sub-populations are replaced. For all fitness and sub-fitness function calculations a fitness score as described before is used. Constraint violations are penalised with a dynamic penalty parameter, which adjusts itself depending on the (sub-)fitness difference between the best and the best feasible agent in each (sub-) population. Full details on this type of weight and how it was calculated can be found in [Smith & Tate 1993] and [Aickelin & Dowsland 2000]. The stopping criterion is the top sub-population showing no improvement for 50 generations.

To obtain statistically sound results all experiments were conducted as 20 runs over all problem instances. All experiments were started with the same set of random seeds, i.e. with the same initial populations. The results are presented in feasibility and cost respectively rent format. Feasibility denotes the probability of finding a feasible solution averaged over all problem instances. Cost / Rent refer to the objective function value of the best feasible solution for each problem instance averaged over the number of instances for which at least one feasible solution was found.

Should the algorithm fail to find a single feasible solution for all 20 runs on one problem instance, a censored observation of one hundred in the nurse case and zero for the mall problem is made instead. As we are minimising the cost for the nurses and maximising the rent of the mall, this is equivalent to a very poor solution. For the nurse-scheduling problem, the cost represents the sum of unfulfilled nurses' requests and unfavourable shift-patterns worked. For the mall, the values for the rent are in thousands of pounds per year.

## 5.2  RESULTS: PARTNERING FOR RECOMBINATION

Table 1 shows the results found by our algorithms for the two problems (N = Nurse problem, M = Mall problem) using the seven different partnering strategies in combination with the pyramidal structure. The results are compared to those found by the standard genetic algorithm (SGA) [Aickelin & Dowsland 2000 and 2001] and the Integer Programming results [Fuller 1998] for the nurse problem and theoretical bounds for the mall problem (both referred to as 'bound'). A number of interesting observations can be made.

In the nurse scheduling case, the SGA approach failed to find good or even feasible solutions for many data sets. This can be explained by the high degree of epistasis present and the inability of the unmodified genetic algorithm to deal with it. Once the pyramidal structure with rank-based selection (S) is introduced, results improve significantly, however there is still room for improvement. For the mall problem, the situation is different. Results found by the SGA are good with high feasibility. This indicates the higher number of feasible solutions for this problem. Solution quality seems reasonably good, too. However, the addition of the pyramidal structure (S) results in a marked deterioration of results.

How can these different results be explained? With the nurse scheduling, the objective function value of a partial solution was obtained by summing the cost values of the nurses and shift-patterns involved. Furthermore, we were able to define relatively meaningful sub-fitness scores by exploiting the 'cumulative' nature of the covering constraints due to the grade structure. Hence, the substitute fitness scores calculated allowed for an effective recombination of partial solutions for the nurse-scheduling problem. Thus, there is a good correlation between the sub-fitness of an agent (and hence its rank and its chance of being selected) and the likelihood that it will form part of a good solution. This also explains why the random (R) scheme produces worse results. The best (B) strategy although giving better results than the random selection fails to solve many problems. However, closer observation of experiments showed that it solved some single data sets well. This indicates that genetic variety is as important as fitness in the evolution of good solutions.

Both the distributed (D) and joint (J) strategies again fail to provide better solutions than the rank-based selection. The distributed strategy is similar to the random strategy as it too ignores fitness scores for selection. Choosing from a fixed pool does have some benefits, as the results are better than for complete random choice. The joint strategy works almost as well as the rank-selection. This shows that the principle of the 'dividing and conquering' works well with the nurse problem split along the grade boundaries. The slightly poorer results can be explained by the 'full' evaluation of all sub-strings although only 'parts' are passed on. Thus, some of the correlation described above is lost.

The two best strategies, both outperforming (S), are partner selection based on attractiveness (A) and choice (C). Again, this further confirms that the partial sub-fitness scores are a good criterion of selection for the pyramidal algorithm. Overall, (C) is better than (A), which corresponds to (C) having a higher selection pressure than (A), which in turn has a higher selection pressure than (S). To conclude, it seems that for this problem a good correlation between agents' sub-fitness, the pyramidal structure and good full solutions exist. Hence, the scheme with the highest selection pressure using most problem specific information scores best.

With the Mall Problem, the situation is more complicated since unlike for the nurse problem a large part of the objective function is a source of epistasis, which the proposed partitioning of the string will not eliminate fully. The constraints are a second source for epistasis. In contrast to the objective function, these depend largely on the whole string, as for instance the total number of shops of a particular size allowed. Only after adding up the shops and sizes for all areas is it known if a solution is feasible or not. So unsurprisingly, a combination of these partial solutions is often unsuccessful because it usually violates the overall constraints.

On their own, solutions of the sub-populations are extremely unlikely to be feasible for the overall problem, as they covered only one fifth of the string. It is equally unlikely for those solutions in the main population, which are formed from the five sub-populations, to be feasible. Although these solutions are of high rent, because the sub-populations ignore the main constraints, their combination is unlikely to produce an overall feasible solution.

The situation is only slightly better with those solutions formed by an individual of the sub-populations and an individual of the main population. Usually, even if the individual of the main population is feasible, the children were not. Again, although the partial string from the sub-population agent was of high rent, it was usually incompatible with the rest of the string, resulting in too many or too few shops of some types. Thus, in contrast to the nurse-scheduling problem, their sub-fitness scores are a far poorer predictor for the compatibility of the parts to form complete solutions.

This is confirmed by the above average performance of the random strategy (R) and the extremely poor results found by the best strategy (B). Similarly to before, the distributed strategy (D) performs well again giving credit to the idea of even selection pressure without relying on fitness scores, whereas the joint strategy (J) performs poorly suffering both from the unsuitable sub-fitness scores and the now hindering pyramidal structure.

Overall, the real winners are again the more complex strategies of choice (C) and attraction (A). At first, this seems contradictory as these rely heavily upon the sub-fitness scores. However, apart from the rank-based initial selection of the first parent, subsequent fitness calculations are made after combining the agents. Since the mall pyramid only has two layers, these combinations are always full solution and hence the full fitness score is used. Thus, the direct link between high fitness and good solutions is re-established. Of the two, (A) performs better than (C). This seems to show that a certain amount of randomness is still important here, which again might be an indication for the lower predictive quality of the sub-fitness scores.

|       | N Cost | N Feasibility | M Rent | M Feasibility |
|-------|--------|---------------|--------|---------------|
| Bound | 8.8    | 100%          | 2640   | 100%          |
| SGA   | 54.2   | 33%           | 1850   | 94%           |
| S     | 17.6   | 75%           | 1540   | 78%           |
| R     | 37.4   | 54%           | 1790   | 86%           |
| B     | 27.1   | 57%           | 1490   | 70%           |
| D     | 26.5   | 61%           | 1770   | 84%           |
| J     | 19.9   | 71%           | 1590   | 78%           |
| A     | 12.2   | 83%           | 1950   | 98%           |
| C     | 11.1   | 87%           | 1910   | 94%           |

Table 1: Partnering Strategies for Recombination Results (N = Nurse, M = Mall).

### 5.3  RESULTS: PARTNERING FOR FITNESS EVALUATION

Table 2 shows the results for a variety of fitness evaluation strategies used and again compares these to the theoretic bounds (Bound) and the standard genetic algorithm approach (SGA). For the Nurse Scheduling Problem all strategies used give better results than those found by the SGA. However, as explained above, most credit for this is attributed to the pyramidal structure reducing epistasis.

On closer examination, rank-based (S), random (R) and distributed (D) perform almost equally well, with the rank-based method being slightly better than the other two. All three methods have in common that they contain a stochastic element in the choice of partner. The benefit of this is apparent when compared to the best (B) method. Here the results are far worse which we attributed to the inherently restricted sampling. Interestingly, using the double schemes improves results across the board, which again strengthens our hypothesis how important good sampling is. The overall best results are found by the double random (RR) method. These results correspond to those reported in [Bull 1997].

The results for the Mall problem are similar to those found for the nurse problem: Double strategies work better than single ones and the Best strategy does particularly poorly. However, unlike for the nurse scheduling none of the single strategies significantly improves results over the SGA approach. Reasons for this have already been outlined in the previous sections, i.e. mainly the nature of splitting the problem into sub-problems being contrary to many of the problem's constraints. On the other hand, even for the simple strategies results are far improved over those found by using the partnering strategies for mating, whilst those found by the double strategies even outperform the SGA. We believe that this can be explained as follows: The main downfall of the partnering for mating strategies for the mall problem was outside those strategies. It lies in the fact that the sub-fitness scores are not a good predictor for the success of sub-solutions. However, as these results show, if the original fitness measures are used combined with good partnering methods the pyramidal structure does work. This confirms our suspicion that the previous 'failure' of the pyramidal idea for the mall problem was rooted within our choice of sub-fitness measure rather than in the hierarchical sub-population idea itself.

|       | N Cost | N Feasibility | M Rent | M Feasibility |
|-------|--------|---------------|--------|---------------|
| Bound | 8.8    | 100%          | 2640   | 100%          |
| SGA   | 54.2   | 33%           | 1850   | 94%           |
| S     | 13.3   | 79%           | 1860   | 90%           |
| R     | 14.5   | 77%           | 1915   | 94%           |
| B     | 35.9   | 44%           | 1550   | 72%           |
| D     | 14.6   | 77%           | 1820   | 88%           |
| SR    | 12.7   | 84%           | 1950   | 99%           |
| BR    | 14.2   | 81%           | 1897   | 86%           |
| RR    | 12.1   | 83%           | 1955   | 99%           |

Table 2: Partnering Strategies for Fitness Evaluation Results (N = Nurse, M = Mall).

## 5.4   RESULTS: NURSE SCHEDULING WITH A HILLCLIMBER

The results presented so far show that even with the best algorithm for the nurse scheduling problem some data instances were unsolvable. In order to overcome this, a special hillclimber has been developed which is fully described in [Aickelin & Dowsland 2001]. The use of local search to refine solutions produced via the GA for complex problem domains is well established – often termed memetic algorithms [e.g. Moscato 1999]. Briefly, the hill-climber is local search based algorithm that iteratively tries to improve solutions by (chain-) swapping shift patterns between nurses or alternatively assigns a strictly solution improving pattern to a nurse. As the hill climber is computationally expensive, it is only used on those solutions showing favourable characteristics for it to exploit. Those solutions are referred to as 'balanced' and one example is a nurse surplus one on day shift and a shortage on another day shift.

The last set of experiments presented in table 3 shows what impact the best partnering schemes for mating (Choice) and evaluation (RR) have once the previously excluded hillclimber is attached to the genetic algorithm. The results reveal that the improvements made by the partnering for mating strategies are equalled by the SGA once both have access to the hill climber. The best solutions are found with the double random fitness evaluation approach coupled with the hill climber. One possible explanation for this effect can be found by having a closer look at the choice mating operator, where an individual picks the best fitting partner from a set. So effectively, a crossover-hill climber strategy is at work here. In the RR case, all gains are also made due to better sampling. However, as mentioned before there is a large stochastic element involved in this case. Judging from these results it seems that again this is beneficial as it leads to a bigger variety of solutions in turn leaving more for the hill climber to exploit.

|  |  | N Cost | N Feasibility |
|---|---|---|---|
| SGA & Hillclimber | SGA&H | 10.8 | 91% |
| Choice & Hillclimber | C&H | 10.7 | 90% |
| RR & Hillclimber | RR&H | 9.9 | 95% |

Table 3: Results for Algorithms combined with a Hillclimber for the Nurse Scheduling Problem.

## 6  CONCLUSIONS

This paper has shown the effect different partner strategies have on a pyramidal coevolutionary genetic algorithm solving two different optimisation problems from the area of multiple-choice scheduling. The recombination results for the five simple strategies (S, R, B, D and J) differ for both problems. This is a reflection of the accurateness of the sub-fitness measure in the sense of its predictive power for sub-solutions to form full solutions following the pyramidal recombination strategies. Therefore, in the case of the nurse problem with a good match between sub-fitness and usefulness for recombination the simple strategies worked well, whereas for the mall problem with its poorer correlation between the two it did not.

The two more advanced strategies (A) and (C) use most problem specific knowledge and work well for both problems. They worked well for the nurse problem because the sub-fitness scores are meaningful. They also worked well for the mall problem because the partners are chosen based on a fitness score after recombination, which in this case equals the full original fitness score. Thus, choosing parents 'post-birth' after evaluating possible children can overcome possible shortcomings in the sub-fitness measure.

Using the partnering strategies for evaluation purposes yields results in accordance with those reported in [Bull 1997]. For both problem the simple strategies worked equally well apart from the restricting 'best' choice. Combining two partnering schemes improved results further with the overall best solutions found by the double random strategy. Interestingly, the improvements of results seemed to be of a different nature than those found by the recombination strategies. There the best method was a hill-climber crossover type approach; here the improvement seems to be based on better sampling and more diversity. Thus for the latter approach an additional hillclimber is able to improve solutions beyond the previously best ones.

## REFERENCES

Ackley, D.H. & Littman, M.L. (1994) Altruism in the Evolution of Communication. In R Brooks & P Maes (eds.) Artificial Life IV, MIT Press, Mass., pp 40-48.

Ahluwalia, M. & Bull, L. (1998) Coevolving Functions in Genetic Programming: Dynamic ADF Creation using GLiB. In V.W. Porto, N. Saravanan, D. Wagen & A.E. Eiben (eds.) Proceedings of the Seventh Annual Conference on Evolutionary Programming. Springer Verlag, pp 809-818.

Aickelin, U. (1999).Genetic Algorithms for Multiple-Choice Optimisation Problems. PhD Dissertation, University of Wales, Swansea, United Kingdom.

Aickelin, U. & Dowsland, K. (2000). Exploiting problem structure in a genetic algorithm approach to a nurse rostering problem. Journal of Scheduling 3, pp 139-153.
|  |  | N Cost | N Feasibility |
|---|---|---|---|
| SGA & Hillclimber | SGA&H | 10.8 | 91% |
| Choice & Hillclimber | C&H | 10.7 | 90% |
| RR & Hillclimber | RR&H | 9.9 | 95% |

Table 3: Results for Algorithms combined with a Hillclimber for the Nurse Scheduling Problem.

## 6  CONCLUSIONS

This paper has shown the effect different partner strategies have on a pyramidal coevolutionary genetic algorithm solving two different optimisation problems from the area of multiple-choice scheduling. The recombination results for the five simple strategies (S, R, B, D and J) differ for both problems. This is a reflection of the accurateness of the sub-fitness measure in the sense of its predictive power for sub-solutions to form full solutions following the pyramidal recombination strategies. Therefore, in the case of the nurse problem with a good match between sub-fitness and usefulness for recombination the simple strategies worked well, whereas for the mall problem with its poorer correlation between the two it did not.

The two more advanced strategies (A) and (C) use most problem specific knowledge and work well for both problems. They worked well for the nurse problem because the sub-fitness scores are meaningful. They also worked well for the mall problem because the partners are chosen based on a fitness score after recombination, which in this case equals the full original fitness score. Thus, choosing parents 'post-birth' after evaluating possible children can overcome possible shortcomings in the sub-fitness measure.

Using the partnering strategies for evaluation purposes yields results in accordance with those reported in [Bull 1997]. For both problem the simple strategies worked equally well apart from the restricting 'best' choice. Combining two partnering schemes improved results further with the overall best solutions found by the double random strategy. Interestingly, the improvements of results seemed to be of a different nature than those found by the recombination strategies. There the best method was a hill-climber crossover type approach; here the improvement seems to be based on better sampling and more diversity. Thus for the latter approach an additional hillclimber is able to improve solutions beyond the previously best ones.

## REFERENCES

Ackley, D.H. & Littman, M.L. (1994) Altruism in the Evolution of Communication. In R Brooks & P Maes (eds.) Artificial Life IV, MIT Press, Mass., pp 40-48.

Ahluwalia, M. & Bull, L. (1998) Coevolving Functions in Genetic Programming: Dynamic ADF Creation using GLiB. In V.W. Porto, N. Saravanan, D. Wagen & A.E. Eiben (eds.) Proceedings of the Seventh Annual Conference on Evolutionary Programming. Springer Verlag, pp 809-818.

Aickelin, U. (1999).Genetic Algorithms for Multiple-Choice Optimisation Problems. PhD Dissertation, University of Wales, Swansea, United Kingdom.

Aickelin, U. & Dowsland, K. (2000). Exploiting problem structure in a genetic algorithm approach to a nurse rostering problem. Journal of Scheduling 3, pp 139-153.

Aickelin, U. & Dowsland, K. (2001). An indirect genetic algorithm approach to a nurse scheduling problem. Under review by the Journal of Computing and Operational Research.

Bean J, Noon C, Ryan S, Salton G (1988) Selecting Tenants in a Shopping Mall. Interfaces 18, pp 1-9.

Bull, L. (1997) Evolutionary Computing in Multi-Agent Environments: Partners. In T. Baeck (ed.) Proceedings of the Seventh International Conference on Genetic Algorithms, Morgan Kaufmann, pp 370-377.

Bull, L. & Fogarty, T.C. (1993) Coevolving Communicating Classifier Systems for Tracking. In R F Albrecht, C R Reeves & N C Steele (eds.) Artificial Neural Networks and Genetic Algorithms, Springer-Verlag, New York, pp 522-527.

Davidor, Y. (1991), Epistasis Variance: A Viewpoint on GA-Hardness. In G Rawlins (ed) Foundations of Genetic Algorithms, Morgan Kaufmann, San Mateo, pp23-35.

Davis, L. (1991) Handbook of Genetic Algorithms. Van Nostrand Reinhold

Fuller, E. (1998) Tackling Scheduling Problems Using Integer Programming. Master Thesis, University of Wales Swansea, United Kingdom.

Holland, J. (1975), Adaptation in Natural and Artificial Systems. Ann Arbor: University of Michigan Press.

Husbands, P. (1994) Distributed Coevolutionary Genetic Algorithms for Multi-Criteria and Multi-Constraint Optimisation. In T.C, Fogarty (ed.) Evolutionary Computing, Springer-Verlag, pp 150-165.

Iba, H. (1996) Emergent Co-operation for Multiple Agents Using Genetic Programming. In H-M Voigt, W Ebeling, I Rechenberg & H-P Schwefel (eds) Parallel Problem Solving from Nature - PPSN IV, Springer, Berlin, pp 32-41.

Koza, J.R. (1991) Genetic Programming. MIT Press.

Martello, S. & Toth, P. (1990), Knapsack Problems, Wiley, Chichester.

Moscato, P. (1999), Memetic Algorithms: A Short Introduction. In D. Corne, M. Dorigo & F. Glover (eds), New Ideas in Optimization, Mc Graw Hill, pp 219-234.

Potter, M. & De Jong, K. (1994) A Co-operative Coevolutionary Approach to Function Optimisation. In Y Davidor, H-P Schwefel & R Manner (eds) Parallel Problem Solving From Nature - PPSN III, Springer-Verlag, Berlin, pp 249-259.

Ronald, E. (1995) When Selection Meets Seduction. In L. Eshelman (ed) Proceedings of the Seventh International Conference on Genetic Algorithms, Morgan Kaufmann, San Francisco, pp 167-173.

Smith, A. & Tate, D. (1993) Genetic Optimisation Using a Penalty. In S. Forrest (ed) Proceedings of the Fifth International Conference on Genetic Algorithms, Morgan Kaufmann, San Francisco, pp 499-505.

Stanley, A. E., Ashlock, D. & Testatsion, L. (1994) Iterated Prisoner's Dilemma with Choice and Refusal of Partners. In C G Langton (ed) Artificial Life III, Addison-Wesley, Redwood City, pp 131-146.

Wolpert, D. & Macready, W. (1997) No Free Lunch Theorem for Search. IEEE Transactions on Evolutionary Computation 1(1), pp 67-82.